\documentclass[copyedit,
ms
]{informs4TD}
\usepackage{graphicx, url}
\usepackage{graphics}
\usepackage{color}
\usepackage{colortbl}  % For colored cells
\usepackage{tablefootnote}
\usepackage{multirow}
\usepackage{array}
\usepackage{mathrsfs}  
\usepackage{amssymb}
\usepackage{amsmath}
\usepackage{natbib}%splitbib,xcolor,appendix,https://www.overleaf.com/project/64982721633797dbfb4dcf03 maybemath,bm,hepparticles,
\usepackage[table,xcdraw]{xcolor}  % 

\usepackage{pgfplots}
\pgfplotsset{compat=1.18}

\usepackage{booktabs}
\usepackage[table]{xcolor}
\usepackage{colortbl}

\definecolor{AIBlue}{RGB}{41,128,185}
\definecolor{OMOrange}{RGB}{211,84,0}
\definecolor{HumanGreen}{RGB}{39,174,96}
\definecolor{FlowPurple}{RGB}{142,68,173}
\definecolor{BackgroundGray}{RGB}{250,251,252}
\definecolor{AccentGold}{RGB}{243,156,18}
\definecolor{DarkGray}{RGB}{52,73,94}

\definecolor{headerblue}{RGB}{45, 62, 80}
\definecolor{rowgray}{RGB}{245, 247, 250}
\definecolor{rowwhite}{RGB}{255, 255, 255}

\usepackage{booktabs}

\usepackage{graphicx}

\TheoremsNumberedThrough

\usepackage{longtable}

\newlength{\defbaselineskip}
\newcommand{\setlinespacing}[1]%
           {\setlength{\baselineskip}{#1 \defbaselineskip}}

\usepackage{natbib}
 \bibpunct[, ]{(}{)}{,}{a}{}{,}%
 \def\bibfont{\small}%

\setcitestyle{citesep={;}}

\makeatletter
\renewcommand\env@matrix[1][c]{\hskip -\arraycolsep
  \let\@ifnextchar\new@ifnextchar
  \array{\c@MaxMatrixCols #1}}
\makeatother

\usepackage{color}
\definecolor{hopkins-blue}{RGB}{0,45,114}
\definecolor{columbia-blue}{RGB}{185, 217, 235}
\definecolor{chicago-maroon}{RGB}{128,0,0}
\definecolor{mit-red}{RGB}{117,0,20}
\definecolor{cornell-red}{RGB}{179,27,27}
\definecolor{lawngreen}{RGB}{0,250,154}
\definecolor{gray}{RGB}{192,192,192}

\newcommand{\look}[1]{#1}

\newif\ifshowinternalnotes
\showinternalnotestrue

\usepackage{accents}

\usepackage{nicefrac}
\usepackage[colorlinks,citecolor=chicago-maroon,urlcolor=chicago-maroon,linkcolor=chicago-maroon]{hyperref}
\usepackage[nameinlink]{cleveref}
\crefname{assumption}{Assumption}{Assumptions}
\crefname{lemma}{Lemma}{Lemmas}
\crefname{theorem}{Theorem}{Theorems}
\crefname{corollary}{Corollary}{Corollaries}
\crefname{proposition}{Proposition}{Propositions}
\crefname{claim}{Claim}{Claims}
\crefname{subclaim}{Subclaim}{Subclaims}
\crefname{procedure}{Procedure}{Procedures}
\crefname{algorithm}{Algorithm}{Algorithms}
\crefname{example}{Example}{Examples}
\crefname{figure}{Figure}{Figures}
\crefname{section}{Section}{Sections}
\crefname{appendix}{Appendix}{Appendices}
\crefname{table}{Table}{Tables}

\definecolor{low}{rgb}{0.9,0.9,0.9}
\definecolor{medium}{rgb}{0.7,0.7,0.9}
\definecolor{high}{rgb}{0.4,0.4,0.8}
\definecolor{keycap}{rgb}{0.9, 0.95, 1}
\definecolor{gray}{rgb}{0.9, 0.95, 0.95}

\usepackage{etoolbox}
\AtBeginDocument{%
  \let\oldbibliography\thebibliography
  \let\endoldbibliography\endthebibliography
  \renewenvironment{thebibliography}[1]{%
    \SingleSpacedXI%
    \oldbibliography{#1}%
  }{%
    \endoldbibliography
  }%
}

\begin{document}
%%%%%%%%%%%%%%%%

\RUNTITLE{Assured Autonomy}

\RUNAUTHOR{Dai, Simchi-Levi, Wu, and Xie}

\TITLE{Assured Autonomy: How Operations Research Powers and Orchestrates Generative AI Systems\thanks{Authors are listed alphabetically.}
}

% Assured Autonomy: How Operations Research Powers and Orchestrates Autonomous AI

\ARTICLEAUTHORS{
\AUTHOR{Tinglong Dai{$^{\dagger,*}$} \hspace{0.25in}
David Simchi-Levi{$^\ddagger$}
\hspace{0.25in}
Michelle Xiao Wu{$^\S$}
\hspace{0.25in}
Yao Xie{$^\P$}}
\smallskip \smallskip
%\thanks{This research received the Johns Hopkins Discovery Award (2015--2016). 
\AFF{
{$^\dagger$}Carey Business School, Johns Hopkins University, Baltimore, Maryland 21202;
Data Science and AI Institute, Johns Hopkins University, Baltimore, Maryland 21218,
\href{mailto:dai@jhu.edu}{\tt dai@jhu.edu}
\\
\smallskip
{$^\ddagger$}Institute for Data, Systems and Society, Operations
Research Center, Department of Civil and Environmental Engineering, Massachusetts Institute of Technology, Cambridge, Massachusetts 02139, \href{mailto:dslevi@mit.edu}{\tt dslevi@mit.edu}
\\
\smallskip
{$^\S$}Purdue University, West Lafayette, Indiana 47907, \href{mailto:michelle.xiao.wu@gmail.com}{\tt michelle.xiao.wu@gmail.com}
\\
\smallskip
{$^\P$}H.\ Milton Stewart School of Industrial and Systems Engineering, Georgia Institute of Technology, Atlanta, Georgia 30332, \href{mailto:yao.xie@isye.gatech.edu}{\tt yao.xie@isye.gatech.edu}
\\
\smallskip
{$^*$}Corresponding author.
}

}
\ABSTRACT{Generative artificial intelligence (GenAI) is shifting from conversational assistants toward agentic systems---autonomous decision-making systems that sense, decide, and act within operational workflows. This shift creates an autonomy paradox: as GenAI systems are granted greater operational autonomy, they should, by design, embody more formal structure, more explicit constraints, and stronger tail-risk discipline. We argue that stochastic generative models can be fragile in operational domains unless paired with mechanisms that provide verifiable feasibility, robustness to distribution shift, and stress testing under high-consequence scenarios. To address this challenge, we develop a conceptual framework for assured autonomy grounded in operations research (OR), built on two complementary approaches. First, \emph{flow-based generative models} frame generation as deterministic transport characterized by an ordinary differential equation, enabling auditability, constraint-aware generation, and connections to optimal transport, robust optimization, and sequential decision control. Second, \emph{operational safety} is formulated through an adversarial robustness lens: decision rules are evaluated against worst-case perturbations within uncertainty or ambiguity sets, making unmodeled risks part of the design. This framework clarifies how increasing autonomy shifts OR's role from solver to guardrail to system architect, with responsibility for control logic, incentive protocols, monitoring regimes, and safety boundaries. These elements define a research agenda for \emph{assured autonomy} in safety-critical, reliability-sensitive operational domains.

\medskip
\noindent\emph{History}: Received January 2026; revised May 2026; accepted May 2026. \emph{Accepting Editor}: Kalyan Singhal.
}


\maketitle

\section{Introduction}\label{sec:intro}

Artificial intelligence (AI) is moving from advice to action. The question is no longer whether generative AI (GenAI) can draft text or write code, but whether an agent can operate---place orders, route vehicles, allocate clinical resources, balance power grids, coordinate logistics---under real constraints and uncertainty. This shift from ``chatbot'' to ``operator'' exposes a paradox that should guide the next decade of Operations Research (OR): greater autonomy demands more structure. We call this the \emph{autonomy paradox}.

\look{To address this paradox, we view assured autonomy as an organizational design problem as much as a modeling one. An autonomous system is credible only when technical choices are paired with clear decision rights and records that make actions inspectable after deployment. In that sense, model quality is necessary, but it is not by itself evidence of safe operation.}\label{sec:intro-bridge}

\look{The design logic is essential because autonomy delivers speed and scale, yet it also amplifies the cost of small errors, hidden constraint violations, and rare failures. In high-stakes settings, expected performance is a weak guide: low-probability regimes can dominate social cost, and ``almost always safe'' is not safe enough.}

\subsection{Operational Risk and the Case for OR}

Even highly engineered autonomous systems exhibit edge cases that trigger recalls, investigations, and public concern. In December 2025, a U.S.\ regulator disclosed a Waymo recall tied to a software issue that could cause vehicles to pass stopped school buses---a rare, high-consequence scenario \citep{WaymoRecallReuters2025}. The lesson generalizes. When AI is embedded in operations, safety depends on engineered discipline: systems auditable and monitorable, respecting hard constraints, and stress-tested against tail events \citep{weick2015}.

Meanwhile, autonomy advances quickly under controlled conditions. A late-2025 lab study shows multi-agent systems built from frontier generative models can manage the Beer Game and reduce total costs relative to human teams \citep{LongSimchiLeviCalmonCalmon2025}. The GenAI Beer Game provides an interactive testbed pairing a natural-language interface with an OR decision engine \citep{Long2025GenAIBeerGame}. Field failures and lab gains point to a single bottleneck: autonomy scales under decision regimes that stay feasible and stable when conditions drift and constraints bind unexpectedly.

These requirements define a design problem: build autonomous operators whose behavior stays feasible, monitorable, and stable under distribution shift and interaction. OR has spent decades developing the corresponding structure---explicit constraints, flow conservation, queueing stability, sequential decision control, and robust planning under uncertainty.

Forged during World War II to orchestrate complex military operations, OR was born in high-stakes settings and carried into infrastructure, logistics, and service systems \citep{Flagle2002,gass_assad_history_or_2011}. Rising autonomy renews that role. In low-autonomy settings, OR acts as a solver. In medium-autonomy settings, it supplies guardrails---constraints, audits, and risk measures. In high-autonomy settings, OR becomes the architect of the operating regime: control logic, incentive protocols, monitoring rules, and safety boundaries within which fleets of agents act.

\subsection{Assured Autonomy as an OR Problem}

A defining feature of assured operational autonomy (``assured autonomy'') is that decisions are sequential, stateful, and coupled over time. Actions taken now reshape future feasibility, risk exposure, and information through delayed and nonlinear dynamics. This distinguishes operational autonomy from one-shot generative tasks and places OR's tradition of sequential decision-making, control, and stability analysis at the center.\footnote{See, e.g., \cite{Powell2023RLSO} and \cite{SuttonBarto2018} for unification perspectives emphasizing closed-loop feasibility and stability rather than reward maximization.} The value of a generative model is not its static realism, but whether the closed-loop system it induces remains feasible, stable, and safe over long horizons.

\look{To formalize scope, we define operational autonomy along six measurable dimensions: action scope ($S$), commitment authority ($C$), action latency ($L$), override structure ($O$), temporal state coupling ($U$), and guardrail tightness ($G$). The pair $(C,O)$ makes delegation boundaries explicit by specifying how decision rights are partitioned between autonomous agents and human supervisors. We define assurance as a closed-loop deployment property requiring feasibility, robustness to shift, detectability, recoverability, and auditability over time. In our architecture, autonomy expands only when each assurance dimension is tied to a measurable signal, a pre-specified control action, and a clearly assigned escalation owner.}\label{sec:def-autonomy-assurance}

The autonomy paradox implies two design commitments. First, the generative mechanism must be constrainable and auditable---closer to an engineered dynamical system than a black-box sampler. This motivates deterministic, flow-based generators---continuous normalizing flows, flow matching, and probability-flow ordinary differential equation (ODE) formulations---that confine randomness to a source distribution while keeping the transformation dynamics deterministic and governed by an ODE. The flow is the engine of generation; transformers provide representations, interfaces, and orchestration.

Second, operational safety should be built through adversarial design rather than post hoc filtering. Catastrophic failures in aviation, supply chains, power grids, or hospitals arise in tail regimes and under interaction; a natural abstraction is game-theoretic. A controller minimizes cost while an adversary---nature, attackers, or distribution shift---maximizes loss, linking assured performance to distributionally robust optimization and to one of OR's historical roots: postwar game theory \citep{gass_assad_history_or_2011,Shubik2002GameTheoryOR}.

These commitments locate OR's leverage. Operational settings expose bottlenecks current generators often sidestep: feasibility by construction, explicit tail regimes that dominate social cost, and generation coupled to optimization so ``plausible'' means decision-impactful rather than surface-similar.

Assured autonomy is autonomy with guarantees: explicit invariants, tail-risk and shift stress tests, and auditable rules for monitoring, escalation, and deferral. It separates systems that look competent in routine settings from those that remain safe when conditions become novel, extreme, or adversarial. Assured autonomy aligns with core OR principles---explicit constraints, worst-case reasoning, and accountable decision rules. Autonomy must be engineered with structure, constraints, and governance, and that redesign is an OR project.

\smallskip
 
This article is conceptual and design-oriented. We specify what assured autonomy requires, diagnose why current GenAI fails under hard constraints and tail risk, and show how OR fills the gap. We synthesize recent work into an OR-powered integration stack---deterministic, transport-style generation; minimax stress testing; optimization and control; and continuous monitoring with fallback---as both blueprint and research agenda.

\look{To make the logic explicit, we organize the paper as a layered architecture for assured autonomy. \cref{sec:gaps} diagnoses operational failure modes in off-the-shelf GenAI; \cref{sec:flows} develops constrained generation as the representation layer; \cref{sec:minimax} develops minimax/DRO stress testing as the robustness layer; \cref{sec:role} formalizes the orchestration layer through monitoring, escalation, fallback, and delegation rights; and Sections~\ref{sec:applications}--\ref{sec:agenda} illustrate and extend the architecture in domain settings. Each layer addresses a distinct failure mode and produces artifacts needed by the next layer.}\label{sec:intro-roadmap}

\section{Why Current GenAI Is Insufficient for Assured Autonomy}
\label{sec:gaps}

GenAI has made rapid gains in producing fluent text and realistic media, and looks compelling in controlled demonstrations. Operational autonomy, however, is judged by a different metric than plausibility. Most GenAI models generate high-probability continuations of observed patterns, whereas operations require actions that satisfy hard constraints, remain stable under feedback and delay, and perform acceptably under distribution shift---especially in rare, correlated regimes that dominate social cost. The failures are \emph{structural}: imitation-based training does not enforce feasibility, stability, or tail robustness. This mismatch sharpens along two dimensions---non-determinism in the decision engine and safety-criticality of the domain. When both are high, residual stochastic error becomes a systematic source of tail risk. Cummings argues ``GenAI is simply too dangerous to include in safety-critical systems'' \citep{Cummings2025ProhibitingGenAIWeaponControl}. The point extends beyond weapons: when rare failure modes cannot be modeled, bounded, and monitored, delegating control to a stochastic generator increases exposure to catastrophic outcomes.

We first explain how stochastic generation complicates certification in safety-critical settings, using large language models (LLMs) and diffusion models as exemplars. We then distinguish semantic from structural constraints that define operational feasibility. Finally, we discuss tail risk, distribution shift, and accountability---the regimes where average-case performance is least informative and post hoc diagnosis is essential.

\subsection{Stochastic Generators and Safety-Critical Control}

LLMs make the gap concrete. An LLM produces the most plausible continuation of a prompt given its training distribution and context. For drafting, summarizing, and brainstorming, this works well. For allocating scarce resources, scheduling tightly coupled activities, or enforcing safety rules that admit no exceptions, it becomes a liability. Correctness here means satisfying feasibility and safety conditions, not producing a reasonable narrative. Experts have argued GenAI should be prohibited ``to control, direct, guide or govern any weapon'' until hallucinations can be modeled and predicted, because the technology remains hard to certify and bound where failures cost lives \citep{Cummings2025ProhibitingGenAIWeaponControl}. Where the acceptable failure rate is effectively zero, plausibility does not substitute for assurance.

Diffusion models \citep{chen2024opportunities,song2020score,song2020denoising} illustrate a related structural misalignment. Standard diffusion destroys structure by adding noise and reconstructs it via a stochastic reverse-time process \citep{anderson1982reverse}---fine for images and text, where intermediate states need not be meaningful. Operational systems differ: feasibility, conservation laws, stability conditions, and safety envelopes must hold along the entire trajectory, not just at the endpoint. A generator that wanders stochastically and is ``corrected'' after the fact is hard to certify and underrepresents tails unless training is redesigned for the rare regimes that dominate social cost.

\subsection{Semantic Versus Structural Constraints}

It is tempting to impose constraints through prompting, penalty terms, or post hoc filtering. But this is brittle: prompts are not enforceable constraints, and soft penalties are not hard feasibility. Even with an external checker, the generator becomes a proposal mechanism whose outputs must be governed by a separate decision regime. Recent work on constrained learning for diffusion models is therefore best read as both progress and diagnosis: Lagrangian-style training can yield satisfaction guarantees for certain constraint classes \citep{khalaficomposition}, yet much of that literature emphasizes semantic or preference-based constraints rather than the conservation, capacity, integrality, and stability constraints that define OR problems.

Distinguishing \emph{semantic} from \emph{structural} constraints helps. Semantic constraints regulate \emph{what} is generated (attributes, labels, preferences, fairness criteria). Structural constraints govern feasibility as a dynamical object---flow conservation, capacity limits, integrality, stability, feasibility over time---and should hold globally throughout execution. Violating a structural constraint produces not a lower-quality output but an infeasible or hard-to-certify action. Assured autonomy thus cannot rest on semantic alignment alone when generative models enter decision loops.

\subsection{Tail Risk, Distribution Shift, and Accountability}

The deeper problem is tail risk and distribution shift. Operational safety is rarely about being right on average; it is about not failing when the system is stressed: when demand surges, disruptions correlate, sensors degrade, or interaction produces congestion and cascades. Standard GenAI objectives reward typical-case fidelity and smooth away rare structures that robust planning should confront. OR frameworks, by contrast, have long emphasized tail events \citep{blanchet2008efficient}, encoding them through chance constraints, ambiguity sets, or worst-case objectives.

Operations also require accountability and diagnosis. High-reliability practice depends on learning from near misses and tracing failures back to their mechanisms \citep{weick2015}. Black-box generators make this difficult: when performance degrades, the cause is often unclear---data drift, prompt drift, hidden constraint violations, or a failure mode absent from training. Without auditable structure, governance becomes reactive and fragile.

\smallskip

\begin{figure}[t]
  \centering
  \includegraphics[width=0.7\linewidth]{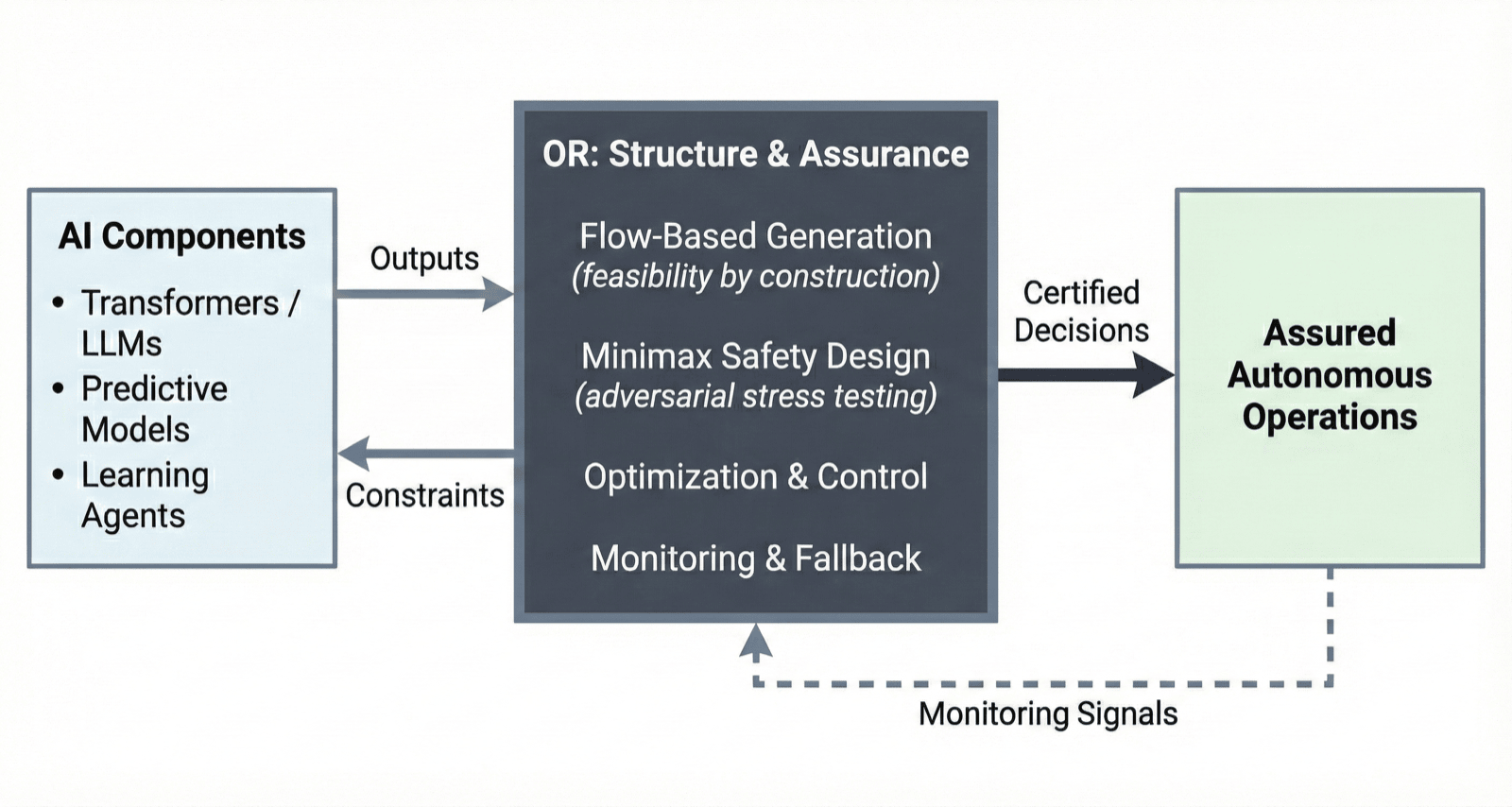}
  \caption{The OR-AI Integration Architecture for Assured Autonomy}
  \label{fig:architecture}
\end{figure}

Off-the-shelf GenAI is not built to enforce trajectory-level feasibility, control tail regimes that dominate social cost, or support accountability in high-reliability operations. These gaps explain why current GenAI remains unreliable as an autonomous operator in safety-critical, constraint-driven settings. The next sections examine technical directions targeting these failure modes, and \cref{fig:architecture} summarizes the OR--AI integration architecture for assured autonomy.

\look{To complement \cref{fig:architecture}, \cref{fig:framework-architecture} maps operational weaknesses (constraint-violation risk, tail-risk blindness, and distribution shift) to autonomy regimes (low, medium, high) and assurance mechanisms. It also organizes key deployment artifacts across layers---constraint-consistent scenarios, feasible decisions, robust-risk certificates, and runtime control signals---and makes their feedback loop explicit so local arguments can be read against the full system design.}

\look{We use $P_0$ to denote the nominal distribution and $P_{\mathrm{adv}}$ to denote the least-favorable distributions for the stress-test. These symbols are used in the minimax and application sections, where $P_{\mathrm{adv}} \in \mathcal{P}(P_0)$ denotes governed distribution shift. We use GenAI for foundation generative models, an agentic system for tool-using decision stacks under guardrails, and an autonomous operator for agents with delegated commitment authority.}\label{sec:notation-glossary}

\begin{figure}[t]
\centering
\begin{tikzpicture}[
  x=1cm,y=1cm,
  font=\small,
  >=stealth,
  line width=0.45pt,
  coltitle/.style={font=\bfseries\small, align=center},
  itembox/.style={
    draw=black!35,
    rounded corners=2pt,
    minimum width=4.1cm,
    minimum height=0.95cm,
    text width=3.75cm,
    align=center,
    inner sep=2.5pt,
    font=\small
  },
  flow/.style={->, black!55, line width=0.5pt},
  feedback/.style={->, black!45, dashed, line width=0.45pt},
]

% ── Column headers ──
\node[coltitle] at (0,3.35) {Operational\\Weaknesses};
\node[coltitle] at (5.2,3.35) {Autonomy\\Regimes};
\node[coltitle] at (10.4,3.35) {Assurance\\Layers};

% ── Column 1: Weaknesses ──
\node[itembox, fill=red!5] (w1) at (0,2.15) {Constraint violations};
\node[itembox, fill=red!5] (w2) at (0,1.00) {Tail-risk blindness};
\node[itembox, fill=red!5] (w3) at (0,-0.15) {Shift fragility};

% ── Column 2: Autonomy regimes ──
\node[itembox, fill=blue!4] (a1) at (5.2,2.15) {Low: advisory};
\node[itembox, fill=blue!4] (a2) at (5.2,1.00) {Medium: propose--certify};
\node[itembox, fill=blue!4] (a3) at (5.2,-0.15) {High: delegated authority};

% ── Column 3: Assurance layers ──
\node[itembox, fill=green!5] (s1) at (10.4,2.15) {Constrained generation};
\node[itembox, fill=green!5] (s2) at (10.4,1.00) {Minimax / DRO stress tests};
\node[itembox, fill=green!5] (s3) at (10.4,-0.15) {Monitoring, escalation, fallback};

% ── Row-aligned arrows ──
\draw[flow] (w1.east) -- (a1.west);
\draw[flow] (w2.east) -- (a2.west);
\draw[flow] (w3.east) -- (a3.west);
\draw[flow] (a1.east) -- (s1.west);
\draw[flow] (a2.east) -- (s2.west);
\draw[flow] (a3.east) -- (s3.west);

% ── Feedback loop ──
\coordinate (fbstart) at ([yshift=-0.20cm]s3.south);
\coordinate (fbend) at ([yshift=-0.20cm]w3.south);
\draw[feedback] (fbstart) .. controls +(0,-0.70) and +(0,-0.70) .. (fbend);
\draw[feedback] (fbend) -- ++(0,0.48);
\node[font=\scriptsize, black!55, align=center] at (5.2,-1.05)
  {Deployment evidence feeds back to weakness diagnosis};

% ── Notation footer ──
\node[font=\scriptsize, black!60, align=center, text width=13.0cm] at (5.2,-1.72) {%
  Nominal: $P_0$ \quad
  Stress test: $P_{\mathrm{adv}} \in \mathcal{P}(P_0)$ \quad
  Artifacts: scenarios, certificates, triggers};

\end{tikzpicture}
\caption{\look{Operational weaknesses, autonomy regimes, and assurance layers. The dashed loop indicates deployment-evidence feedback. Rows indicate primary emphasis, not exclusive pairings; all assurance layers may operate across autonomy regimes, with their relative importance increasing as delegated authority expands.}}
\label{fig:framework-architecture}
\end{figure}

\section{Flow-Based Generative Models: Structured Evolution in Probability Space}
\label{sec:flows}

We introduce flow-based generative models as a design pattern for auditable, constrainable generation---not as a claim of empirical superiority. Flow-based models construct complex distributions by transporting probability mass through a sequence of maps, from discrete normalizing flows to continuous-time formulations based on neural ODEs and transport-based variants \citep{Dinh2017,rezende2015variational,albergo2023stochastic,chen2018neural,lipman2023,geng2025mean,song2023consistency,xu2023normalizing}. The transport formulation is implementation-agnostic: simple or kernel-based maps suffice in low-dimensional settings, while neural parameterizations matter when high-dimensional heterogeneity makes expressivity the bottleneck \citep{peyre2019computational}. Through an OR lens, flow-based generation is an iterative algorithm in probability space \citep{xie2025flow}, making it natural to impose constraints and risk functionals on distributional evolution. The rest of this section formalizes continuous-time distributional dynamics, shows how deterministic transport yields structure-by-construction and auditability, and explains how the interface supports constraint- and tail-aware scenario generation.

\look{In \cref{fig:framework-architecture}, this section corresponds to the representation layer: turning raw model capability into constraint-aware, auditable scenario generation that can be governed downstream.}

\subsection{Continuous-Time Dynamics and Distributional Evolution}

In the continuous-time setting, let $X_t \in \mathbb{R}^d$ denote the system state at time $t \in [0,T]$, with initial condition
\(
X_0 \sim \rho_0,
\)
where $\rho_0$ is a source distribution (a simple reference such as Gaussian in generative use, or the empirical data distribution in analytical use; see below). A flow-based model specifies a time-dependent velocity field
\(
v_\phi : \mathbb{R}^d \times [0,T] \to \mathbb{R}^d,
\)
parameterized by $\phi$, and the system evolves according to the ordinary differential equation
\begin{equation}
\frac{d X_t}{d t} = v_\phi(X_t,t), \quad t \in [0, T].
\label{eq:flow_ode}
\end{equation}
As particles evolve under these deterministic dynamics, the induced probability distribution evolves continuously. Let $\rho_t(x)$ denote the probability density of $X_t$. Its evolution is governed by the continuity equation
\begin{equation}
\partial_t \rho_t(x) + \nabla \cdot \big( \rho_t(x)\, v_\phi(x,t) \big) = 0,
\label{eq:continuity_equation}
\end{equation}
which expresses conservation of probability mass along the flow.

Under mild regularity conditions, the dynamics in~\cref{eq:flow_ode} are deterministic and invertible, yielding explicit transport between distributions. The velocity field $v_\phi$ is typically a neural network---often residual or time-incremental---learned via variational objectives in probability space or flow-matching formulations. Gaussian references are common, but the framework is not restricted: the terminal distribution may represent structured or adversarial targets, such as least-favorable distributions in distributionally robust optimization \citep{xu2024flow}.

The same transport admits two interpretations. In generative use, one draws samples from a reference distribution $Q$ and pushes them forward to obtain samples from a target distribution $P$. In analytical or decision-oriented settings, the flow pushes data or scenarios from $P$ toward a worst-case $Q$. Both viewpoints rely on the same transport structure and differ only in how the learned dynamics are traversed. This connects to Wasserstein gradient-flow structure and JKO-type constructions \citep{ambrosio2005gradient,cheng2024convergence,jordan1998variational}, providing a variational view of training and stability aligned with OR's optimization foundations.

From an OR perspective, \cref{eq:flow_ode,eq:continuity_equation} make a central duality explicit. Constraints can be imposed on probability measures---e.g., requiring $\rho_t$ to concentrate mass on a feasible set or control tail risk---or on individual sample paths, by designing $v_\phi(\cdot,t)$ so trajectories preserve invariants or remain within a feasible region. Once the velocity field is fixed, both particle paths and induced distributional evolution are determined, yielding feasibility by construction.

\begin{figure}[t]
\centering
\includegraphics[width=1\linewidth]{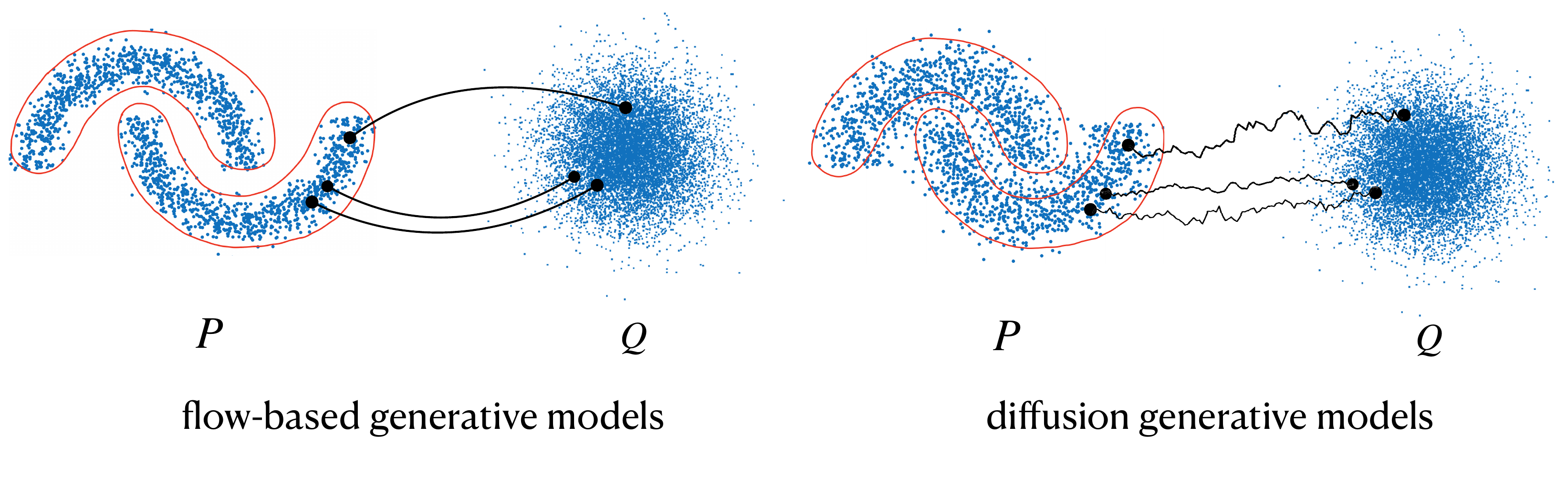}
\caption{
Structural contrast between flow-based and diffusion-based generative models under constraints.
Red curves denote a constrained feasible region. Left: flow-based models transport samples deterministically from a reference distribution \(Q\) to a target distribution \(P\), allowing constraints to be incorporated directly into the transformation. Right: diffusion-based models generate samples through iterative stochastic refinement, with randomness injected at each step and constraints typically enforced through guidance or correction mechanisms.
}
\label{fig:flow-model}
\end{figure}

Diffusion-based generators can also incorporate constraints, typically through stochastic sampling with guidance, projection, or correction steps. These mechanisms work well in many applications, but make hard sample-wise certification and traceability harder. \cref{fig:flow-model} illustrates the contrast: \look{deterministic-by-design transport makes auditability, replayability, and enforcement interfaces intrinsic, and these features matter disproportionately in safety-critical operational loops.}

\subsection{Deterministic Transport and Structure-by-Construction}

Modern flow-based generative modeling can be understood as learning transport maps between probability distributions. Rather than treating uncertainty as static input, this perspective models system evolution as a structured transformation of distributions governed by invariants, feasibility conditions, and stability requirements. In OR, many constraints are identities, not preferences---flow conservation, capacity limits, nonnegativity, balance conditions, integrality, or temporal ordering---that hold globally. A transport-based view accommodates these: distributions evolve under dynamics designed to preserve structure by construction, rather than enforced post hoc through rejection, penalties, or repair. Generation becomes a controlled process governed by structure, not a black-box sampling procedure aimed at reproducing observed data.

Diffusion and score-based models can be interpreted within this framework but typically realize it through stochastic dynamics. Diffusion models generate samples by simulating noisy reverse-time processes \citep{ho2020denoising,song2020score}, and \look{LLMs often introduce randomness during sampling-based decoding; under fixed decoding regimes they can be reproducible, but reproducibility alone does not enforce operational feasibility} \citep{brown2020language}. Stochasticity enables diversity in creative domains. In operations, it becomes a liability: inconsistent outputs, intermittent feasibility violations, and failure modes that resist diagnosis. While diffusion models admit a deterministic sampling formulation via the probability-flow ODE under idealized conditions \citep{song2020score}, their predominant formulations emphasize stochastic sampling, with determinism playing a secondary role.

Flow-based generative models align with the transport-map view. Randomness is confined to the initial draw; generation then proceeds through deterministic dynamics governed by an ODE. This enables exact traceability and replayability: an output can be traced to a specific initial state, its atypicality quantified, and the trajectory replayed for auditing. Operationally, the deterministic transport map becomes the central object of assured autonomy, exposing the control surfaces required for feasibility enforcement, monitoring, and governance. Recent developments---flow matching, consistency models, and mean-flow formulations---decouple transport from density estimation, easing computational concerns while preserving determinism and controllability for safety-critical decision-making \citep{geng2025mean,lipman2023,song2023consistency}.

\look{Deterministic transport improves replayability, debugging, and audit tracing because identical initial states and fixed solvers produce identical trajectories. By itself, however, determinism does not guarantee safety or certifiability; it can also scale misspecification faster if invariants are wrong. Assurance still depends on invariant specification, robustness to distribution shift, numerically stable integration, and runtime monitoring with fallback authority. Diffusion models can also admit deterministic probability-flow formulations; our claim is therefore architectural rather than model-family exclusive. Determinism contributes to assurance only when coupled with explicit constraints, stress testing, and governance controls.}\label{sec:determinism-calibration}

Operational deployments rarely require training from scratch. A practical alternative is to start from a pretrained generative or representation model and adapt it via fine-tuning or lightweight updates---conditioning layers, adapters, or low-rank parameterizations \citep{hu2022lora}---using limited in-domain data and operational objectives. In the transport-map view, this warm-starts the velocity field and refines it to encode domain constraints, tail-sensitive regimes, and decision-coupled loss signals, reducing data and compute requirements while improving reliability. This is where OR enters: constraint penalties, risk functionals, and decision-layer feedback can be imposed during fine-tuning rather than deferred to inference-time correction.

\section{Game-Theoretic Safety and Robust Autonomy}
\label{sec:minimax}

\begin{figure}[t]
\centering
\includegraphics[width=0.6\linewidth]{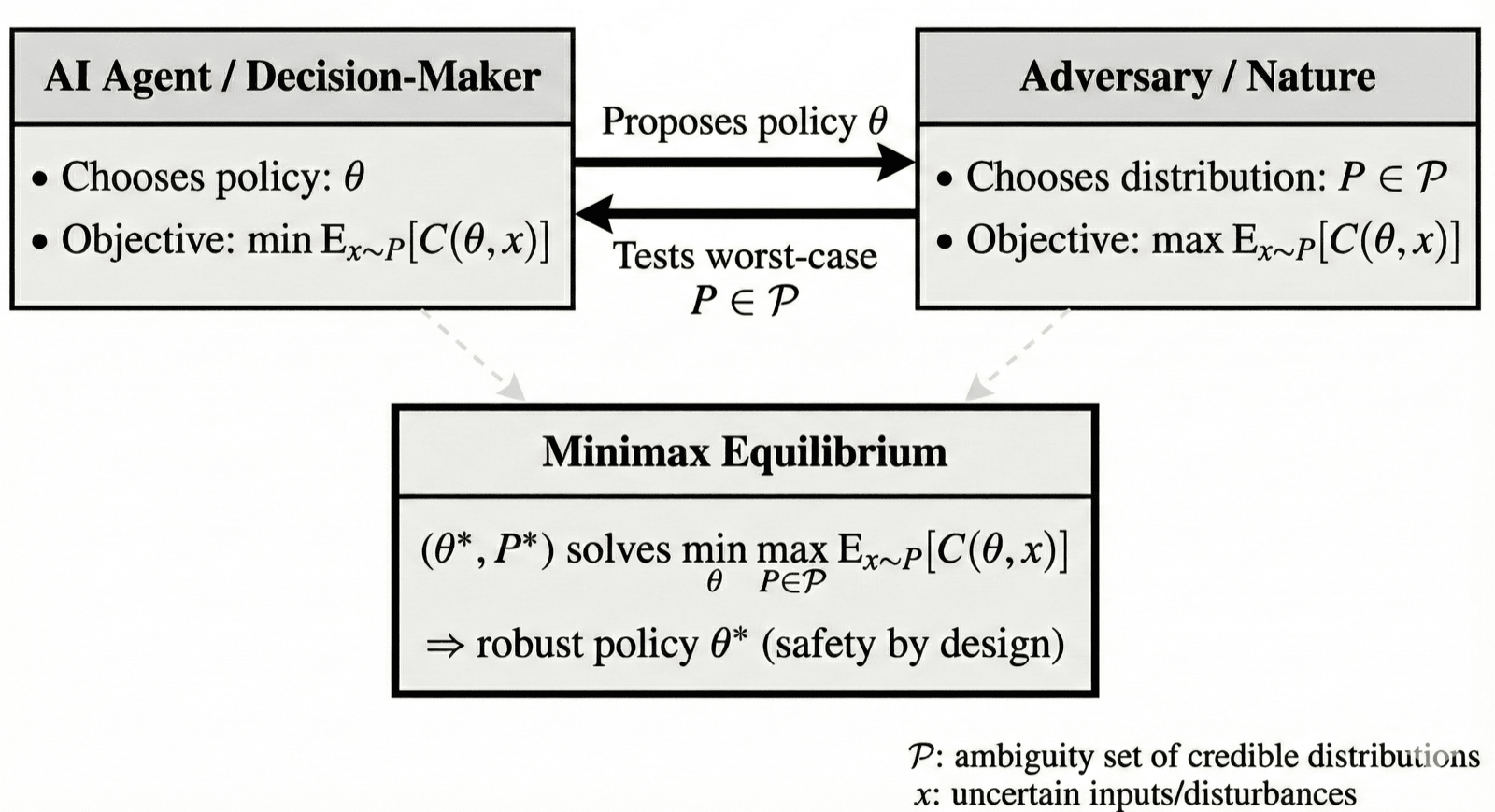}
\caption{Minimax Game-Theoretic Framework for AI Safety}
\label{fig:minimax}
\end{figure}

Constrainability is not safety. A generator that faithfully reproduces the training distribution still fails when demand spikes, sensors degrade, or an adversary probes for weaknesses. Assured autonomy requires stress testing---systematic confrontation with futures that have not yet occurred. OR formalizes this as a game: the decision-maker chooses a policy; an adversary chooses the scenario that breaks it. Operational autonomy is ultimately a confrontation with surprise. Models improve and sensors proliferate, yet the world produces combinations outside yesterday's training distribution. If an autonomous system is judged by what happens when things go wrong, safety should be designed for those regimes, not appended afterward. OR has a name for this stance: robustness, formalized as a game between a decision-maker and an adversary representing nature, strategic opponents, or distribution shift. \cref{fig:minimax} previews this logic: the designer chooses a policy, an adversary selects a worst-case shift within a credible ambiguity set, and safety is the resulting equilibrium, not a post-hoc patch.

\look{In \cref{fig:framework-architecture}, this is the robustness layer: stress-testing nominal plans under least-favorable shifts before autonomy is delegated and decision rights are expanded.}

A coupling between generative models and decisions is the stochastic optimization problem
\begin{equation}
\min_{\theta}\ \mathbb{E}_{x \sim P_0}\big[C(\theta,x)\big],
\label{eq:baseline_decision}
\end{equation}
where $\theta$ denotes a decision or controller, $x$ represents realized disturbances and inputs, $C(\theta,x)$ is the induced operational loss, and $P_0$ is the nominal designer distribution learned or estimated from data. This captures the classical ``average-case'' objective but leaves systems exposed to rare, correlated, or unmodeled regimes that dominate operational risk.

A canonical robust formulation is distributionally robust optimization (DRO) \citep[see, e.g.,][]{blanchet2019quantifying,gao2023distributionally,kuhn2019wasserstein,selvi2025differential,shapiro2017distributionally,wang2025sinkhorn}:
\begin{equation}\label{eq:minimax_general}
\min_{\theta} \max_{P_{\mathrm{adv}} \in \mathcal{P}(P_0)} \; \mathbb{E}_{x \sim P_{\mathrm{adv}}}\!\left[C(\theta,x)\right],
\end{equation}
where $\mathcal{P}(P_0)$ is an ambiguity set of plausible shifts around the nominal designer distribution $P_0$. The designer chooses $\theta$; the adversary chooses $P_{\mathrm{adv}}$; the resulting equilibrium trades off performance against vulnerability. \cref{eq:minimax_general} makes ``what could go wrong?'' operational: specify how the environment may shift, then optimize against the least favorable case.

In the GenAI setting, related minimax and worst-case formulations have emerged in recent work on high-dimensional problems \citep{ChengXieZhuZhu2025WorstCase,xu2024flow}. These developments reflect growing recognition that flow-based generative models can represent and learn complex high-dimensional worst-case distributions, enabling direct sample generation from such adversarial models.

\subsection{Distribution-Level Constraints and Sample-Level Enforcement}

Operational constraint control distinguishes two classes.
\emph{Semantic constraints} regulate output attributes or preferences; \emph{structural constraints} govern feasibility as a dynamical object---flow conservation, capacity limits, integrality, or stability---and should hold globally.
In OR, structural constraints are non-negotiable identities, not preferences.

For generative models used in decision-making, this distinction motivates imposing constraints on the \emph{generated scenarios}, and hence on the induced scenario distribution.
Let $x \in \mathcal{X}$ denote an operational scenario (e.g., demand trajectories, lead times, outages), drawn from $P$, and let $h_k(x) \le 0$ encode structural feasibility requirements.
When generation is coupled to downstream decisions, a natural objective is
\[
\mathcal{J}(P) = \min_{\theta \in \Theta} \mathbb{E}_{x \sim P}[C(\theta,x)],
\]
which links constrained scenario generation directly to the minimax formulation in~\cref{eq:minimax_general}.

A canonical formulation selects (or learns) a distribution $P$ within an admissible model class $\mathcal{P}$ such that feasibility holds with high probability:
\begin{equation}
\max_{P \in \mathcal{P}}\ \mathcal{J}(P)
\quad \text{s.t.} \quad
\mathbb{P}_{x \sim P}\!\big(h_k(x) \le 0\big) \ge 1-\varepsilon_k,
\quad k=1,\ldots,K.
\label{eq:gen_chance_constraints_p}
\end{equation}
More generally, feasibility can be enforced through risk-sensitive functionals,
\begin{equation}
\max_{P \in \mathcal{P}}\ \mathcal{J}(P)
\quad \text{s.t.} \quad
\mathbb{E}_{x \sim P}\!\big[\psi_k(h_k(x))\big] \le \delta_k,
\quad k=1,\ldots,K,
\label{eq:gen_risk_constraints_p}
\end{equation}
where $\psi_k(\cdot)$ encodes tail-weighted or risk-sensitive penalties such as hinge losses or Conditional Value at Risk (CVaR)-type functions.
Lagrange multipliers yield a relaxation over \emph{distribution-level} constraint violations,
\begin{equation}
\min_{P \in \mathcal{P}} \ \max_{\lambda_k \ge 0}~~
-\mathcal{J}(P)
+ \sum_{k=1}^K \lambda_k
\left(
\mathbb{E}_{x \sim P}\!\big[\psi_k(h_k(x))\big] - \delta_k
\right),
\label{eq:gen_lagrangian_p}
\end{equation}
which highlights an OR-specific distinction: $h_k$ represent \emph{global structural feasibility}, not semantic attributes, and feasibility is enforced over generated operational scenarios rather than model parameters.

Flow-based generators have a practical advantage: distribution-level requirements can be realized at the sample level. Because flow models implement generation as a deterministic transport map, pointwise constraints---upper bounds, conservation laws, state-dependent safety conditions---can be enforced or penalized along generated trajectories. Constraints on the induced distribution can often be implemented by shaping the transport dynamics. This sample--distribution duality is natural in OR, where feasibility is defined pointwise rather than over model parameters, and it aligns with stochastic programming.

Comparable constraint control is possible in diffusion-based generators, typically through stochastic dynamics with guidance, projection, or correction during sampling. Such approaches can make hard sample-wise guarantees and traceability more delicate. The explicit transport structure of flow-based models makes constraint handling transparent and directly compatible with OR formulations of feasibility and risk.

\look{Integrality-heavy applications require an explicit continuous-to-discrete bridge. For routing, assignment, batching, and capacity-commitment decisions, we use a hybrid design in which continuous generation proposes structured scenarios while a mixed-integer/combinatorial layer enforces discrete feasibility at execution. Actions requiring projection or repair are logged as assurance events; repeated repairs are treated as monitoring warnings that can tighten admissible action sets or trigger human review. In practice, integrality constraints define hard governance boundaries, not soft preferences.}\label{sec:discrete-combinatorial}

\subsection{Minimax Stress Testing and Least-Favorable Distributions}

Minimax stress testing treats reliability as performance against an adversary. The idea traces to von Neumann's zero-sum games \citep{vonNeumann1944} and enters OR through robust optimization and robust control, converting worst-case logic into tractable prescriptions \citep{BenTalNemirovski2002Robust}. Distributionally robust optimization (DRO) places ambiguity on the data-generating process, often through moment sets, $\phi$-divergences, or Wasserstein neighborhoods \citep{Rahimian2019}. This provides a calibration problem: which stress classes are credible, and how much probability mass should represent misspecification?

Two implications are central. First, robustness targets more than overt attacks. Operational disasters concentrate in rare, correlated, cascading regimes that dominate social cost. High-reliability organizations institutionalize a sustained ``preoccupation with failure'' \citep{weick2015}. Grid operators formalize it with contingency standards; aviation relies on certification culture; hospitals operationalize escalation protocols. Across domains, expected performance is a weak standard. The operational standard is avoiding catastrophic modes within defined stress classes.

Second, minimax reasoning becomes engineering only when the adversary is computable. Worst-case design requires a searchable representation. Deterministic generators supply one. When uncertainty is modeled as a transport map pushing a reference measure into operational scenarios, the adversary in \cref{eq:minimax_general} can be parameterized by that map. The problem becomes worst-case \emph{generation}: the adversary explores a continuous family of distributions in $\mathcal{P}(P_0)$ rather than a menu of hand-built stress tests. Recent work characterizing least-favorable distributions in Wasserstein space as pushforwards of transport maps provides this bridge \citep{ChengXieZhuZhu2025WorstCase}. In data-scarce regimes, the generator need only represent the ambiguity set with structure and control; fidelity to the true process is secondary. This yields decision-coupled stress testing: the generator searches for failures damaging to the deployed policy $\theta$, not generic extremes.

\look{For implementation, we recommend a six-step robust-design protocol: define invariants and mission loss; identify plausible shift classes; choose ambiguity geometry (e.g., Wasserstein, divergence-based, moment-based, or event-wise); calibrate uncertainty size on historical stress windows; tune the conservatism--performance frontier; and bind runtime triggers to robust-bound violations. Reporting nominal and robust outcomes jointly keeps the conservatism dial explicit for managerial and regulatory review and forces transparent trade-offs between false reassurance and over-conservatism.}\label{sec:dro-design-recipe}

\subsection{Interaction, Governance, and Operational Escalation}

Robust autonomy has an interaction dimension. Many operational settings are systems of agents---vehicles negotiating right-of-way, inventory nodes coordinating replenishment, software agents bidding in markets. Safety depends on incentives, equilibrium selection, and protocol design alongside worst-case uncertainty. OR's game-theoretic toolkit is central: design communication and commitment rules that prevent deadlock, align local objectives with system goals, and rule out pathological equilibria. The aim is to reduce the strategic degrees of freedom through which interaction produces systemic harm.

Robustness without governance is incomplete. Deployment is not solving \cref{eq:minimax_general} once; drift, nonstationarity, and updates make control ongoing. Assured autonomy needs an explicit monitoring-and-deference policy: what is measured, what counts as ``out of control,'' and what action follows. OR's sequential decision and monitoring traditions apply: escalation is itself a control policy, and fallback is a designed operating regime, not an improvised human response.

\section{From Solver to System Architect: The Evolving Role of OR in Assured Autonomy}
\label{sec:role}

OR began by solving well-posed decision problems. Humans supplied judgment and executed the plan. With partial automation, OR shifted to assurance: checking feasibility, bounding actions, auditing recommendations from heuristics or learning systems. As operational autonomy rises, that separation no longer holds. OR must design the decision regime---the rules, constraints, monitoring triggers, escalation policies, and coordination protocols governing how agents act and interact. \look{Within the layered architecture introduced in \cref{sec:intro-roadmap}, this section is the orchestration layer that turns constrained generation (\cref{sec:flows}) and minimax stress testing (\cref{sec:minimax}) into deployable assurance.}\label{sec:architecture-roadmap}

\cref{tab:evolution} summarizes the change. In decision support, AI is predictive and OR prescriptive. Optimization turns forecasts and risk scores into plans, and a human decides. In partial autonomy, the stack becomes ``propose-then-certify.'' A learning component proposes; OR certifies against hard constraints and risk limits, projects onto the feasible set as needed, and triggers overrides or reversion to conservative policies when monitoring signals elevated risk.

\begin{table}[t]
\centering
\caption{Evolution of OR Roles with Increasing AI Autonomy.}
\label{tab:evolution}
\small
\renewcommand{\arraystretch}{1.35}
\setlength{\tabcolsep}{8pt}
\begin{tabular}{@{}>{\raggedright\arraybackslash}p{3.8cm}
                >{\raggedright\arraybackslash}p{4.5cm}
                >{\raggedright\arraybackslash}p{5.5cm}@{}}
\toprule
\textit{Level of autonomy} & \textit{AI paradigm} & \textit{Role of OR} \\
\midrule
Decision support (human in charge) &
Predictive analytics &
Solver: provides optimal or near-optimal solutions \\[1.5ex]
Human-in-the-loop automation &
Stochastic generative AI &
Guardrail: sets constraints, validates AI suggestions \\[1.5ex]
Fully autonomous operations &
\look{Constraint-aware, auditable generative AI} &
Architect/legislator: designs system, rules, and objectives \\
\bottomrule
\end{tabular}
\end{table}

\subsection{Designing Decision Regimes for Autonomous Operations}

In fully autonomous settings---a dark factory, an autonomous supply chain, or a smart city where signals and vehicles negotiate flows in real time---the OR task is constitutional. OR specifies the regime: admissible actions, objectives, inviolable constraints, information rights, and conflict-resolution procedures. The design object is no longer an instance, but the system producing instances and acting on them.

This shift is visible in power systems. Operators once used optimal power flow to recommend adjustments humans executed. As renewables and fast-acting grid-edge devices proliferate, control becomes automated, and OR's emphasis moves to rules and limits: reserve requirements, frequency-response obligations, and operating envelopes that maintain reliability. OR practitioners write the ``grid code'' autonomous controllers must satisfy.

A parallel shift appears in automated fulfillment. In robotized warehouses, managers do not route individual robots. OR defines the operating regime: zoning policies, priority rules, congestion management, and conflict-resolution logic that prevents deadlock while preserving throughput. Optimization remains, but shapes protocols rather than selecting each move.

OR fits this role because architecture is a choice under constraints. Autonomous systems need rules that are efficient, robust, and aligned with system objectives. Mechanism design offers an analogy: choose rules so decentralized behavior yields acceptable outcomes. In autonomy, the ``players'' are software modules. The designer decides what each module controls, observes, and how conflicts are resolved.

\subsection{Monitoring, Escalation, and Fallback}

A second task is fallback. Every autonomous system meets regimes outside its validated envelope, and assured autonomy requires explicit fallback rules: when monitoring signals out-of-control behavior, the system contracts its action space, reverts to conservative policies, or defers to humans. OR defines the triggers and optimizes fallback behavior so safety is preserved while residual performance is retained.

\look{Operationally, monitoring should be implemented as a signal-trigger-action authority loop. Representative signals include safety-envelope slack, detection statistics, repeated constraint-binding counts, queue-instability indicators, and near-miss precursor rates. Hard-threshold breaches tighten admissible actions; persistent breaches trigger conservative policy reversion; and sustained high-risk states require human authorization before commitment. Thresholds should be calibrated as an explicit operating trade-off among missed detections, false alarms, detection delays, and human-review workload. Each escalation action should be logged with state, trigger, action, and rationale to preserve auditability.}\label{sec:runtime-monitoring-fallback}

\look{The same pattern applies to autonomous model-debugging workflows in OR. Recent solver-in-the-loop benchmarks study self-correction workflows in which an agent uses solver feedback (including irreducible-infeasible-subset (IIS) diagnostics) to identify and repair infeasibilities iteratively \citep{ao2026solverintheloopmdpbasedbenchmarksselfcorrection}. Infeasibility serves as an escalation signal, one that comes with a traceable path back to resolution.}

The legislator's role includes compliance. External requirements---regulation, safety standards, organizational policy---must translate into executable constraints. Fairness requirements in hiring or credit are increasingly mathematical constraints. For high-stakes autonomy, compliance comes by design when constraints and auditability are built in.

This shift implies meta-level optimization: the upper level chooses objectives, loss functions, information structures, and coordination protocols; the lower-level outcome is the behavior induced by learning and interaction. The bi-level task is to choose rules so the equilibrium is stable and aligned.

\subsection{OR Inside the GenAI Stack: Inference and Resource Scheduling}

The OR--GenAI interaction is bidirectional. OR supplies structure and assurance when generative models enter operational loops; OR also improves the GenAI stack itself. Inference scheduling, memory and compute allocation, and latency--throughput tradeoffs in LLM serving are queueing and scheduling problems. Recent work shows OR models can guide online LLM inference scheduling under memory constraints \citep{AoLuoSimchiLeviWang2025}. OR governs both the decision regime and the computational substrate on which agentic AI runs.

\subsection{Implications for OR Research and Practice}

Supply chain autonomy illustrates this. In Beer-Game testbeds, multi-agent GenAI systems can outperform human teams, yet performance and stability hinge on the regime---what information is shared, what constraints cap extreme actions, how costs are defined \citep{LongSimchiLeviCalmonCalmon2025,SimchiLeviDaiMenacheWu2025}. The GenAI Beer Game makes the regime explicit by embedding generative agents in a structured operational system, enabling study of auditability, constraint adherence, and failure modes beyond static benchmarks \citep{Long2025GenAIBeerGame}. As autonomy scales, OR moves up the ladder---from solving instances, to certifying actions, to designing the regime---calling for a broader toolkit treating learning, adaptation, and multi-agent interaction as design objects.

\section{Applications Across Domains}\label{sec:applications}
The preceding sections argued that operational autonomy scales only when \emph{assured} by design. The layered architecture introduced in \cref{sec:intro-roadmap}---constrained generation, minimax stress testing, and orchestration---defines the design problem; this section illustrates what it implies in practice, where the autonomy paradox binds most tightly, and how OR's role evolves as autonomy increases (\cref{tab:domain_map}). \look{Relative to \cref{fig:framework-architecture}, this section instantiates the full stack in concrete sectors and shows how regime-specific invariants, escalation rules, and delegation boundaries differ in deployment.}

\begin{table}[t]
\centering
\caption{Assured Autonomy Across Domains: What Becomes Autonomous, What Should Never Be Violated, and Where OR Provides the ``Assurance'' Layer.}
\label{tab:domain_map}
\small
\renewcommand{\arraystretch}{1.35}
\setlength{\tabcolsep}{5pt}
\begin{tabular}{@{}>{\raggedright\arraybackslash}p{0.12\linewidth} 
                >{\raggedright\arraybackslash}p{0.22\linewidth} 
                >{\raggedright\arraybackslash}p{0.24\linewidth} 
                >{\raggedright\arraybackslash}p{0.34\linewidth}@{}}
\toprule
\textit{Domain} & \textit{Operational autonomy} & \textit{Invariants \& tail risks} & \textit{OR assurance mechanisms} \\
\midrule
Supply chains & 
Multi-agent replenishment, allocation, and routing over rolling horizons & 
Flow balance, capacity, lead times; cascades (bullwhip) under shocks and delays & 
Network optimization and robust control; constraint-preserving scenario generation; adversarial stress tests \\[1.5ex]
Mobility \& aviation & 
Real-time trajectory choice with interaction among many agents & 
Right-of-way and separation minima; rare sensor/comm failures; interaction tail events & 
Constrained optimal control and reachability; certified safety envelopes; worst-case interaction/weather generators \\[1.5ex]
Healthcare operations & 
Workflow autonomy (documentation, triage, scheduling) and selective clinical support & 
``Never-miss'' events; case-mix shift; propagation of errors through workflows & 
Queueing and resource allocation with safety constraints; statistical monitoring; explicit deferral and escalation rules \\[1.5ex]
Power grids & 
Automated dispatch, reserve management, and remedial actions at machine speed & 
N-1 / N-1-1 security and stability; correlated outages; extreme weather & 
Security-constrained OPF/UC; contingency analysis; DRO/minimax over outage distributions; real-time monitoring \\
\bottomrule
\end{tabular}
\end{table}

\subsection{Supply Chains}\label{subsec:supplychains}

Supply chains are a clean laboratory for the autonomy paradox: locally sensible actions, under delay and uncertainty, can generate globally unstable dynamics. The bullwhip effect formalizes this amplification analytically and experimentally in the Beer Distribution Game tradition \citep{LeePadmanabhanWhang1997,Sterman1989}. As autonomy rises, the design object shifts from average cost reduction to closed-loop stability: will a multi-agent system remain well behaved when information is inconsistent, conditions shift, or one agent makes a rare extreme mistake?

Inventory planning is the canonical OR setting for sequential decision-making under uncertainty. Long before modern reinforcement learning, dynamic programming and stochastic inventory theory produced practical policy classes---base-stock and $(s,S)$ policies---offering a transparent lens on service levels, stability, and tail-risk trade-offs \citep{Porteus2002,Zipkin2000}. In an assured-autonomy framing, these policies define admissible control structures, provide interpretable fallback modes, and anchor evaluation when learning-based components are introduced.

The interface layer changes faster than the optimization layer. As LLMs reduce the cost of specifying objectives, constraints, and ``what-if'' analyses, the limiting factor becomes the reliability of the optimization primitives. Early deployments use LLMs as a natural-language layer translating planner intent into OR-based optimization and simulation workflows \citep{MenachePathuriSimchiLeviLinton2025,SimchiLeviMellouMenachePathuri2025}, accelerating formulation rather than replacing solvers \citep{SimchiLeviDaiMenacheWu2025}. Autonomy reallocates OR effort toward admissible action spaces, invariants, and stress tests inside digital twins.

Recent evidence shows promise and limits. Multi-agent supply chains can be operated autonomously in simulation using frontier generative models \citep{LongSimchiLeviCalmonCalmon2025}. Yet gains hinge less on ``free reasoning'' than on OR design choices: explicit objectives, information policies, and hard constraints preventing destabilizing actions (e.g., budget caps or action bounds blocking panic ordering). \look{In this evidence-to-deployment bridge, LLM-agent simulations are policy-evidence generators, not policy executors. Before simulated outputs inform real decisions, they should pass a constraints-first gate: feasibility screening, distribution-shift stress testing, and pre-specified escalation/fallback logic. Because human and LLM behaviors can diverge out of distribution, delegated decision rights should remain conditional on runtime monitoring and override readiness, not model confidence alone.}\label{sec:llm-agent-simulation} Autonomy arrives when the supply chain is treated as a controlled dynamical system with admissible inputs.

Flow-based generation and minimax safety make this operational. Supply chain uncertainty spans demand, lead times, yield, capacity, and disruptions, so generators should preserve inventory balance, nonnegativity, and capacity. A deterministic flow maps a latent source into constraint-consistent scenarios, making generation auditable inside the decision loop. The minimax layer enforces tail resilience via \cref{eq:minimax_general}: choose $\theta$ against $P_{\mathrm{adv}} \in \mathcal{P}(P_0)$ over demand-and-disruption paths $x$ with cost $C(\theta,x)$. What is new is computable worst cases: in Wasserstein formulations, least-favorable distributions arise as transported measures, giving a continuous adversary and decision-coupled stress tests rather than a fixed scenario menu \citep{ChengXieZhuZhu2025WorstCase,xu2024flow}.

\look{A compact two-echelon inventory example illustrates the full stack end-to-end. Let plant inventory be $I_t^P$, distribution-center on-hand inventory be $I_t^D$, backlog be $B_t$, production be $x_t$, shipment be $y_t$, and arrival to the distribution center be $a_t$. The exogenous uncertainty is $\xi=(D_{1:T},L_{1:T},K_{1:T})$, where $K_t$ is plant capacity, $D_t$ is stochastic demand, and $L_t$ is shipment lead time. A learned constrained generative model produces $\xi=G_\phi(z;\omega)$, where $z$ is latent noise, $\omega$ encodes operational context, and $\phi$ denotes generator parameters; the model is trained so that generated demand, lead-time, and capacity paths preserve nonnegativity, temporal dependence, and admissible capacity envelopes. This induces a nominal distribution $P_0$ over admissible scenario paths. Given $P_0$, the decision layer chooses a causal replenishment policy $\pi\in\Pi$ that maps observed history into production and shipment decisions. For simplicity, treat $L_t$ as an integer lead time realized when the shipment leaves the plant. The resulting pathwise dynamics satisfy
\[
I_{t+1}^P = I_t^P + x_t - y_t,\qquad 0 \le x_t \le K_t,\qquad 0 \le y_t \le I_t^P + x_t,
\]
\[
a_t = \sum_{u=1}^t y_u\,\mathbf 1\{u+L_u=t\},\qquad
I_{t+1}^D - B_{t+1} = I_t^D - B_t + a_t - D_t,
\]
\[
I_{t+1}^D \ge 0,\qquad B_{t+1} \ge 0, \qquad I_{t+1}^D B_{t+1} = 0.
\]
A stylized minimax formulation is
\[
\min_{\pi\in\Pi}\ \sup_{P_{\mathrm{adv}}\in\mathcal P(P_0)}
\mathbb{E}_{\xi\sim P_{\mathrm{adv}}}\!\left[\sum_{t=1}^{T}\bigl(c x_t + h_P I_{t+1}^P + h_D I_{t+1}^D + p B_{t+1}\bigr)\right],
\]
where $\mathcal{P}(P_0)$ is an ambiguity set around $P_0$, for example a Wasserstein, divergence-based, or moment-based neighborhood. The service target may be imposed through a worst-case stockout constraint,
\[
\sup_{P_{\mathrm{adv}}\in\mathcal P(P_0)} \mathbb{P}_{P_{\mathrm{adv}}}(B_{t+1}>0)\le 1-\alpha,\qquad t=1,\ldots,T,
\]
or through a worst-case tail-risk bound such as
\[
\sup_{P_{\mathrm{adv}}\in\mathcal P(P_0)} \operatorname{CVaR}_{\beta}(B_{t+1})\le \tau.
\]
In deployment, runtime monitoring tracks service slippage, repeated capacity binding, lead-time spikes, and bullwhip proxies such as rolling upstream replenishment variance relative to demand variance. Threshold breaches trigger conservative base-stock fallback; persistent breaches require human authorization under a pre-specified authority matrix.}\label{sec:worked-example-inventory}

Monitoring must be designed in. Key signals are dynamical: rising upstream order variance, inventory oscillations, repeated constraint binding. Statistical process control is built for these patterns, except the ``process'' now includes autonomous agents whose behavior can drift with prompts, context windows, and upstream data. A Six-Sigma approach makes this operational: control charts for decision stability (e.g., bullwhip proxies) and escalation rules triggering conservative modes or human review when the system leaves control \citep{Montgomery2019SPC}. Because supply chains evolve over hours to weeks, adversarial stress tests can run continuously in a digital twin, enabling intervention before instability turns catastrophic.

\subsection{Mobility and Aviation}\label{subsec:aviation}

Transportation autonomy is often framed as a perception breakthrough. In practice, the binding constraints are coordination and regulation. The standard is compliance with bright-line safety rules where other users behave strategically, unpredictably, or simply incorrectly. When an autonomous vehicle mishandles a stopped school bus, regulators treat it as a rule violation, not a marginal performance miss---hence recalls and scrutiny \citep{WaymoRecallReuters2025}. When robotaxi programs stumble, enforcement turns on incident response, reporting, and verifiable risk controls \citep{NHTSA2024}.

These features sharpen the autonomy paradox. Mobility agents act in shared space. If the generative core is stochastic---e.g., iterative denoising---it is hard to certify that separation, stopping, and right-of-way constraints hold along the entire trajectory rather than on average or after repair. Flow-based generation shifts the burden from cleanup to design. When trajectories arise from deterministic dynamics with a constraint-aware vector field, feasibility becomes an invariant of motion. This is OR's comparative advantage: specify admissible dynamics rather than projecting infeasible samples back.

Aviation shows what assured autonomy looks like in a mature safety regime. The U.S.\ National Airspace System relies on decision support for time-based traffic management that schedules and meters flows through constrained airspace \citep{FAA2024}. The logic is OR by construction: demand--capacity balancing, network constraints, scheduling with separation requirements. Generative or agentic components must remain subordinate to certified invariants. Transformers can help with intent inference, coordination messages, and explanations. The trajectory engine and supervisory layer guaranteeing separation must be auditable, stress-testable, and compatible with certification practice.

The minimax layer is equally concrete. The adversary is the coupled process creating conflict: weather restricting airspace, surveillance or communication degradation, and interaction patterns producing dense encounters. An ambiguity set can bound joint shifts in weather, demand, and sensing error. Transport-based generators offer a tractable representation, and minimax optimization supplies the forcing function: a controller safe against least-favorable distributions inside the ambiguity set supports a stronger safety case.

The key distinction from supply chains is timescale and certification. Mobility and aviation run at machine speed (milliseconds to minutes) under tight regulatory norms. Monitoring must therefore be continuous, fast, and action-guiding. In mobility, leading indicators include near-miss rates, envelope violations, and repeated rule conflicts; in aviation, loss-of-separation precursors and sustained overload in constrained sectors. Escalation should be explicit: tighten the envelope, revert to conservative policies, or hand off when signals indicate loss of control.

\subsection{Healthcare Operations}\label{subsec:healthcare}

Healthcare is where the autonomy debate most often blurs categories. ``Workflow autonomy''---drafting notes, coding visits, summarizing encounters---differs from ``clinical autonomy,'' where errors can harm patients. Early GenAI gains reflect this: ambient scribes reduce documentation burden and burnout \citep{Dai2025}, while even documentation automation raises governance questions about consent and trust \citep{JAMAOpenInformedConsentAmbientAI2025}. The lesson is visible at low stakes: deployment succeeds only with design, monitoring, and accountability.

Assurance becomes decisive when systems touch operations: emergency-department triage, inpatient bed assignment, operating-room scheduling. Here invariants are clinical and operational at once: avoid catastrophic misses, prevent unsafe delays, preserve stability during surges. Tail risk dominates because capacity is finite. Small shifts in arrivals or acuity push the system past a threshold where queues grow rapidly and errors propagate across units. ``AI checks AI'' is a weak safeguard. A second model often shares the same data and blind spots and does not encode hospital constraints. Assurance belongs outside the language model: statistical monitoring for drift in calibration and case mix, with OR decision logic that tightens admissible actions, triggers review, or reverts to conservative policies when signals indicate loss of control.

Flow-based generative models fit this assurance layer because they generate the right object: structured stress tests rather than text. Hospitals need plausible surge paths, correlated resource shortfalls, and patient-flow trajectories respecting capacity, conservation, and time ordering. Transport-based generators can produce such trajectories while preserving feasibility by construction. A minimax formulation turns stress testing into a design principle: an adversary searches within an ambiguity set for arrival, acuity, and service-time distributions that strain the system most, and the decision maker chooses staffing, bed allocation, and triage thresholds safe under least-favorable conditions.

The domain implication is selective autonomy. Administrative work and low-risk allocations can be automated; high-risk decisions require hard boundaries via formal deferral rules, triggered by monitoring statistics and calibrated to the cost of false reassurance. Assured autonomy in healthcare protects clinical attention for cases where autonomy is least appropriate.

\subsection{Power Grids}\label{subsec:powergrids}

If any sector already lives under the logic of assured autonomy, it is the power grid. Operators do not optimize ``on average.'' They operate under explicit invariants: supply equals demand, flows respect thermal and voltage limits, and the system remains stable under credible contingencies. These are codified in reliability standards compelling performance under defined classes of adverse events, including sequential contingencies and stability constraints \citep{NERCTPL00151}. Regional planning turns this mandate into routine practice through contingency analysis and reinforcement rules; N-1-1 studies apply a first contingency, allow prescribed corrective actions, then impose a second contingency \citep{PJMManual14B}. This is minimax reasoning embedded in institutions: plan for adverse regimes, not typical days.

Agentic AI changes the surface of grid operations but not the governing principle. Forecasting, anomaly detection, and remedial-action recommendation can become more autonomous, increasing speed and complexity in a coupled cyber-physical system. The autonomy paradox follows: faster control demands a decision regime that is deterministic, certifiable, and constraint-preserving by construction. This is why flow-based generative modeling fits the grid. Power systems are organized around flows and dynamics; uncertain inputs---renewable ramps, weather-driven demand, correlated outages---can be represented as a deterministic transport map from a latent source to physically plausible scenario distributions respecting domain structure.

The minimax layer becomes concrete. The adversary is nature and correlation: extreme weather, common-mode failures, contingency sequences pushing toward cascading collapse. An ambiguity set can encode plausible distribution shift via Wasserstein neighborhoods around empirical distributions of weather and renewable generation. The planner chooses reserves, dispatch, and corrective actions that perform well against least-favorable distributions, consistent with the reliability paradigm. Recent work on worst-case generation in Wasserstein space constructs least-favorable distributions as pushforwards of a transport map rather than from a finite catalog of hand-crafted scenarios \citep{ChengXieZhuZhu2025WorstCase}.

The grid clarifies a governance lesson: monitoring and fallback are part of design. Power systems already have telemetry, alarms, and protection logic. Assured autonomy means aligning AI components with that discipline. Recommendations should be filtered through security constraints; deviations detected using established stability and adequacy metrics; fallback made explicit, from conservative redispatch to automated protection and, in extremis, controlled load shedding. The point is not to replace the reliability-first regime but to make agentic AI compatible with it: transformers improve interface and explanation, while deterministic flows and explicit optimization preserve invariants the grid cannot compromise.

\medskip

Across the four domains, the lesson is the same. Autonomy fails when operations demand invariants, tail-risk discipline, and coordination that stochastic generation and after-the-fact checks cannot supply. Assured autonomy rests on three choices: \look{an auditable, constraint-aware generator, often implemented through deterministic transport}; minimax stress tests searching for catastrophic regimes; and monitoring with clear escalation and fallback rules. The binding constraint differs by domain---delays and feedback in supply chains, certified interaction in mobility and aviation, workflow propagation in healthcare, codified reliability under correlated contingencies in grids. The agenda is to move assurance upstream: invariants in the model, worst cases in the loop, escalation in the system.

\section{Research Agenda}
\label{sec:agenda}

Assured autonomy requires treating autonomous systems as engineered objects: feasible by construction, robust to distribution shift and tail events, and governable over their lifecycle. Four priorities follow: (i) feasibility by construction at the learning--optimization interface; (ii) minimax safety, worst-case generation, and verification; (iii) monitoring, handoffs, and lifecycle governance; (iv) public goods that make progress cumulative.

\subsection{Feasibility by Construction: Learning Meets Optimization}

Autonomy collapses the separation between learning and optimization: the model is not merely input to a solver; the solver is part of the model. The priority is making feasibility and invariants native to learning, not imposed afterward. One route embeds optimization primitives within learning pipelines via differentiable optimization layers, enabling end-to-end training with explicit constraints \citep{AmosKolter2017OptNet,DontiAmosKolter2017TaskBasedLearning,WilderDilysKang2019Melding}. A second develops sequential decision methods maintaining safety at each step, not only on average. Constrained Markov decision processes provide the right abstraction, with the frontier in scaling constraint satisfaction under function approximation and partial observability \citep{Altman1999CMDP,GarciaFernandez2015SafeRL}. The key object is closed-loop behavior: does the learned policy preserve feasibility, stability, and conservation under the shifts and interactions deployment induces? Progress is testable with operational metrics---constraint-violation rates, long-horizon stability, stress-regime performance---rather than prediction error or episodic reward alone.

\subsection{Minimax Safety, Worst-Case Generation, and Verification}

Assured autonomy requires treating rare disasters as design inputs. This calls for objectives optimizing worst-case performance over credible ambiguity sets, connecting to robust optimization and DRO \citep{BenTalNemirovski1998Robust,BenTalNemirovski2002Robust,EsfahaniKuhn2018WassersteinDRO,Rahimian2019}. The minimax template provides a unifying lens: the decision-maker selects a policy while an adversary selects stressors or distributions exposing failure regimes. The challenge is tractability at scale. Recent work on worst-case generation over Wasserstein space suggests one path: represent least-favorable distributions as pushforwards of transport maps, yielding continuous worst-case generators \citep{ChengXieZhuZhu2025WorstCase}. Related developments linking flow-based learning to equilibrium computation in mean-field systems point to a broader bridge between transport, control, and multi-agent autonomy \citep{YuLeeXieCheng2025MFGFlowMatching}.

Verification is equally central. It should answer concrete questions: can the system enter an unsafe state; can it violate a hard constraint; can plausible perturbations trigger cascading failure? Runtime assurance and safety-filtering ideas from control---control barrier functions and shielding---enforce hard constraints online while learning-based components operate within provably safe regions \citep{Alshiekh2018,Ames2017}. Neural network verification and formal methods provide additional building blocks \citep{KatzBarrettDillJulianKochenderfer2017Reluplex,TjengXiaoTedrake2019Evaluating}. OR has a comparative advantage because many such tasks reduce to constrained counterexample search, where mixed-integer optimization and robust search are natural \citep{FischettiJo2018MIPVerify}.

\subsection{Monitoring, Handoffs, and Lifecycle Governance}

Even well-designed autonomous systems drift: data distributions change, incentives evolve, correlated failures emerge at scale. Assured autonomy depends as much on governance as on algorithms. Monitoring should be a control function detecting when the system leaves its specification and triggering predefined responses. A useful template is Statistical Process Control (SPC): continuous measurement of performance and violation metrics, calibrated alarms, and explicit escalation rules \citep{Montgomery2019SPC,Shewhart1931EconomicControl}. At its core is a sequential detection problem---identifying change-points or distributional shifts quickly \citep{xie2021sequential}---emphasizing interpretability and operational relevance.

Beyond detection, handoffs are the central object. Escalation and fallback decisions trade off delay against false alarms and shape the system's safety--performance profile \citep{weick2015}. The handoff rule itself is a policy to optimize: what triggers deferral, what information is transferred, and how the system learns from overrides without creating new failure modes---subject to safety, workload, and delay constraints when multiple agents and humans share a workflow.

\subsection{Benchmarks and Data Readiness}

A final constraint is infrastructural. Progress needs benchmarks small enough to iterate on yet structured enough to capture operational reality. Vision and language advanced when shared datasets made progress legible \citep{Deng2009ImageNet}; RL benefited from shared environments standardizing comparison \citep{Brockman2016OpenAIGym}. Assured autonomy needs analogous testbeds for constraint-preserving generation, tail-aware stress testing, and decision quality under shift. Metrics should be operational---regret, stability, violation rates, worst-case performance---not likelihood alone.

Digital twins offer a related opportunity, but only if structured as stress-testing environments generating rare events, shifts, and worst-case scenarios under feasibility constraints. Without that structure, twins reinforce average-case behavior rather than surface the failure modes governing safety. The data-readiness gap is the binding complement: operational datasets are abundant but rarely curated to expose constraint sets, invariants, regime shifts, and rare-event structure. Reusable datasets and testbeds are the basis for cumulative science and credible external validity.

\smallskip

What ties these priorities together is a vision of autonomy: feasible by construction, hardened by minimax stress testing, governed by explicit monitoring and handoff rules, accelerated by public benchmarks. Meeting that standard is the price of scaling autonomy responsibly, and OR's opportunity to shape what autonomy becomes.

\section{Conclusion}
\label{sec:concl}
\label{sec:conclusion}

Autonomous AI is leaving the lab and entering operations---warehouses, supply chains, clinics, mobility, and finance. The gains are real; the failures are fast and sticky. As autonomy rises, structure matters more. OR will shift from optimizing within a given workflow to designing the workflow itself.

This pattern echoes earlier waves of automation. Societies did not make high-energy machines acceptable by trusting average-case performance; they embedded control, standards, and contingency procedures in the surrounding system. Financial markets became resilient not because trading algorithms grew ``smarter'' but because trading operated within risk limits, monitoring, and circuit breakers. Autonomy faces the same test. The relevant test is not whether a model can generate plausible outputs, but whether the operational system preserves nonnegotiable invariants, remains stable in tail regimes, and switches to a safe mode when conditions leave the validated envelope. Methodologically, this reframes autonomy as controlled evolution in probability space, with constraints and tail risk treated as explicit design requirements.

The practical implication: investing in models alone will disappoint. Value comes from redesigning the decision regime around the model---defining admissible actions, specifying escalation and deference rules, and institutionalizing stress testing and monitoring as operational processes. The managerial task is to decide \emph{where} autonomy is appropriate, \emph{how} it is bounded, and \emph{when} it should yield to conservative policies or human judgment.

The research agenda is similarly concrete. Many central questions are not solved by scale: how to enforce feasibility throughout generation; how to search systematically for consequential failure regimes; how to verify safety properties without relying on model self-assessment; and how to coordinate multiple agents with aligned incentives and stable dynamics. These problems sit naturally in the OR toolkit, echoing Herbert Simon's argument that AI and OR are stronger together than apart \citep{Simon1987}.

The autonomy paradox organizes this paper. Autonomy increases the value of rules that can be audited and enforced. ``Assured autonomy'' treats constraints, tail regimes, and monitoring as design inputs, not afterthoughts. With that discipline, autonomy scales; without it, autonomy imports unpriced risk. It is not a natural extension of current GenAI pipelines. It must be engineered---and OR provides the language and tools.

\look{Assured autonomy is a system property achieved jointly through constrained generation, adversarial robustness, and runtime governance; no single component is sufficient in isolation. The OR contribution is therefore architectural: designing the coupled regime of constraints, stress tests, delegation rights, and escalation policies that keep autonomy productive while safe under shift and interaction.}

\ACKNOWLEDGMENT{The work of Y. Xie is partially supported by NSF DMS-2134037, CMMI-2112533, and the Coca-Cola Foundation.}

\bigskip

%\SingleSpacedXI
\bibliography{GenAI_refs}
\bibliographystyle{informs2014}
%%%%%%%%%%%%%%%%%
\end{document}